\pdfoutput=1

\documentclass[11pt]{article}

\usepackage[preprint]{acl}

\usepackage{times}
\usepackage{latexsym}
\usepackage{graphicx}
\usepackage{array}
\usepackage{tabularx}
\usepackage{float}
\usepackage{tabularx,colortbl}
\usepackage{multirow}
\usepackage{comment}
\usepackage{booktabs}    
\usepackage{caption} 
\usepackage{hyperref} 

\usepackage[T1]{fontenc}

\usepackage[utf8]{inputenc}

\usepackage{microtype}

\usepackage{inconsolata}

\title{Zero-Shot Learning and Key Points Are All You Need for Automated Fact-Checking}

\author{
	\textbf{Mohammad Ghiasvand Mohammadkhani$^{1}$ \quad Ali Ghiasvand Mohammadkhani$^{2}$ \quad Hamid Beigy$^{3}$} \\
	$^{1}$Amirkabir University of Technology \quad $^{2}$Shahid Soltani 4 High School \quad $^{3}$Sharif University of Technology \\
	\texttt{mohammad.ghiasvand@aut.ac.ir} \quad \texttt{aghiasvandm@gmail.com} \quad \texttt{beigy@sharif.edu}
}

\begin{document}

\maketitle

\begin{abstract}
Automated fact-checking is an important task because determining the accurate status of a proposed claim within the vast amount of information available online is a critical challenge. This challenge requires robust evaluation to prevent the spread of false information. Modern large language models (LLMs) have demonstrated high capability in performing a diverse range of Natural Language Processing (NLP) tasks. By utilizing proper prompting strategies, their versatility—due to their understanding of large context sizes and zero-shot learning ability—enables them to simulate human problem-solving intuition and move towards being an alternative to humans for solving problems. In this work, we introduce a straightforward framework based on \textit{\textbf{Z}ero-\textbf{S}hot \textbf{L}earning} and \textit{{\textbf{Ke}y \textbf{P}oints}} (ZSL-KeP) for automated fact-checking, which despite its simplicity, performed well on the AVeriTeC shared task dataset by robustly improving the baseline and achieving 10\textsuperscript{th} place.\footnote{\footnotesize{Code and data released at  \url{https://github.com/mghiasvand1/ZSL-KeP}}}
\end{abstract}

\section{Introduction}
The AVeriTeC task \cite{schlichtkrull2024averitec} is designed to encourage the development of advanced frameworks for automated fact-checking, a critical task in NLP. With the rapid spread of information and misinformation online, automated fact-checking is increasingly important. Given the time-consuming nature of manual fact-checking, building an effective neural language model-based framework is valuable for saving time and costs, improving performance, and supporting human judgment. Significant efforts are being made to automate this process within digital tools or LLMs \cite{nakov2021automated}.

LLMs with billions of parameters offer extensive knowledge and strong reasoning capabilities that can be customized for various tasks. Designing effective and appropriate prompts is crucial in this customization process. Recent utilization of LLMs can mainly be divided into two categories: fine-tuning and In-Context Learning (ICL). Given the enormous size of LLMs and the high computational cost associated with fine-tuning them, utilizing ICL through zero-shot or few-shot prompting is much more efficient. 

Explaining the reasoning behind a decision is crucial for user trust in automated fact-checking, as users need to understand the evidence behind the model's verdict \cite{guo2022survey}. This work employs Large Language Models (LLMs) with Zero-Shot Learning (ZSL), which offer advantages over simpler, classification-based models due to their long context windows and high reasoning capabilities. Besides using powerful LLMs and effective prompting, accurate retrieval of relevant information is vital. This involves hierarchical, step-by-step prompting and decomposition-based retrieval methods \cite{zhang2023towards}. This paper describes the novel approach implemented by our team, \textbf{MA-Bros-H}, for the AVeriTeC shared task, which integrates ZSL and key point utilization within a unified and straightforward framework.

\section{Related Works}

\begin{figure*}[h]
	\centering
	\includegraphics[width=0.94\textwidth]{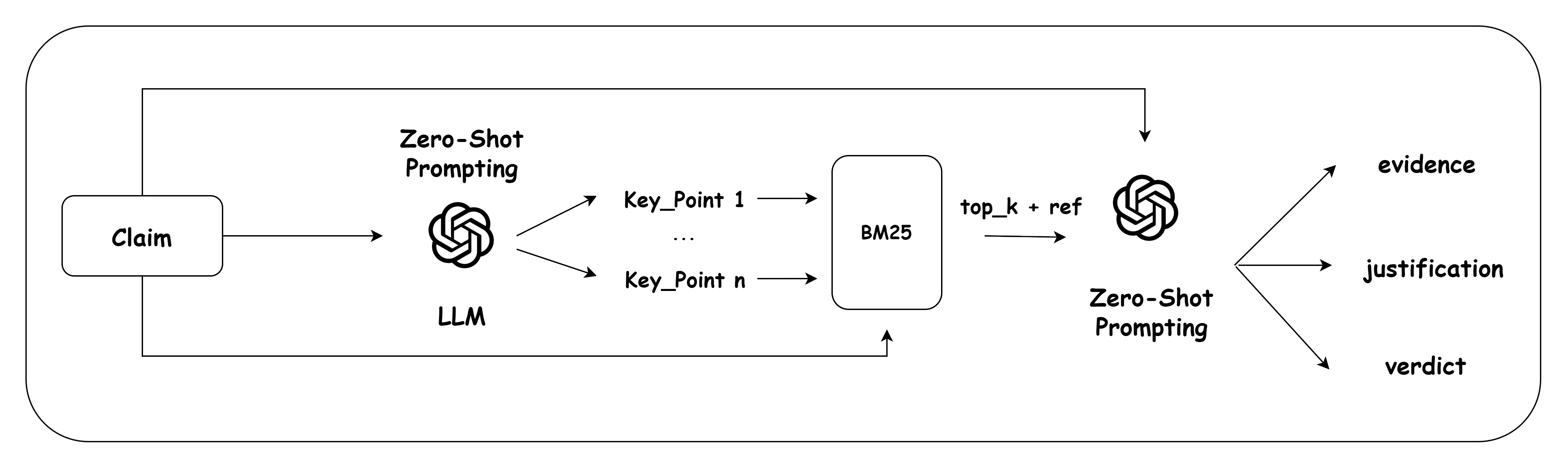}
	\caption{ZSL-KeP Framework Illustration}
	\label{fig:model_arch}
\end{figure*}

To highlight a few recent research efforts in automated fact-checking, it is notable that \cite{kotonya2020explainable} provided explainability through summarization, and \cite{lee2020language} utilized the internal knowledge of pretrained language models such as BERT \cite{devlin2018bert} within their framework. Additionally, \cite{lee2021towards} employed few-shot prompting for fact-checking, \cite{zhang2023towards} introduced a hierarchical, step-by-step prompting method that involves claim decomposition followed by step-by-step reasoning to predict the final verdict, and \cite{kim2024can} proposed a multi-agent debate strategy for explainable fact-checking.

\section{Methodology} 

\begin{table*}[htbp]
	\captionsetup{font=footnotesize}
	\centering
	\begin{tabular}{l|rr|c}
		\toprule
		Method & Q only & Q + A & AVeriTeC score \\ \midrule
		AVeriTeC Baseline \cite{schlichtkrull2024averitec} & 0.24 & 0.20 & 0.11\\ \midrule
		ZSL-KeP (Ours) & 0.38 & 0.24 & 0.27\\
		\bottomrule
	\end{tabular}
	\caption{Main results include retrieval scores for both questions alone and for questions with answers, as well as the AVeriTeC score for the baseline and our proposed method.}
	\label{table:results}
\end{table*}

This section provides a detailed overview of the problem definition of AVeriTeC task, as well as the operational procedure of our ZSL-KeP model, as outlined in Figure \ref{fig:model_arch}.

\subsection{Problem Definition}
Since our method is explicitly based on zero-shot prompting, we use only the test data to execute our framework, ignoring the train and validation datasets. For each data point in the test dataset, a claim is provided, and a verdict must be predicted from the labels “\textit{Supported}”, “\textit{Refuted}”, “\textit{Not Enough Evidence}”, and “\textit{Conflicting Evidence/Cherry-Picking}”. Additionally, for each claim, a JSON file called a knowledge store is provided. This file contains numerous URLs with scraped texts, including some gold documents that assist in selecting the accurate label. The expected output includes a verdict for the input claim and adequate, yet non-redundant, evidence, preferably in the form of question-answer pairs, along with the corresponding URL and scraped text for each pair to justify the source of each proposed question-answer pair. It is noteworthy that the answer type for each question can be “\textit{Extractive}”, “\textit{Abstractive}”, “\textit{Boolean}” or “\textit{Unanswerable}”.

\subsection{ZSL-KeP Framework}
Our ZSL-KeP framework is a procedure that contains multiple steps detailed below. However, compared to the baseline method proposed in \cite{schlichtkrull2024averitec}, our method is much more straightforward, containing fewer steps than the baseline, does not require any fine-tuning, and is simpler to implement.

\subsubsection{Zero-Shot Key Points Construction}
\label{firsts}
In the first step, we receive the claim as input and aim to construct key points based on the received claim using ZSL with our chosen LLM. The primary objective of forming key points is that even a simple claim can contain several key points. When searching and retrieving information from the knowledge store, more extensive retrieval typically yields more comprehensive information. A claim might not return many helpful documents when queried directly, but by constructing diverse key points from it, we can obtain more relevant and diverse information. As shown in the prompting template in Appendix \ref{ap:pt}, we limit the number of primitive key points to four. For these distinct key points, we ask the LLM to identify and return pairs of key points whose combinations result in valuable and richer key points. This process aims to construct an extensive set of key points based on the input claim, facilitating more divergent retrieval in the next step. 

\subsubsection{Extensive Retrieval with References}
As mentioned, for each claim, we have a large knowledge store consisting of various URLs with their scraped texts, among which the gold documents for selecting the best and correct verdict are present. In the previous step, we constructed several key points for each claim, either of a normal type or paired, as explained earlier. If the number of constructed key points is \(n\), we treat these key points as a list of queries. We append the main input claim to this list and use BM25 \cite{robertson2009probabilistic} to retrieve results for each of the \(n+1\) queries with a different {\textit{top}\_\textit{k}} parameter for each query. For each selected retrieval result, since each JSON file contains many URLs and each URL has several scraped texts, we construct an ID by concatenating the URL index within the JSON file with an underscore, followed by the index of the scraped text within the list. For each retrieval document, we attach the text “<ID>” (where ID is the constructed corresponding ID) to the document. After retrieving and appending all these documents for each query, we separate them with a newline character. Finally, we concatenate all groups of retrievals, separating them with two newline characters and several dashes in between, to form a unified retrieval string for the input claim.

\subsubsection{Zero-Shot Prediction}
\label{thirds}
In this stage, which is the final step of our framework, we use ZSL to generate evidence, followed by a justification and, finally, a verdict. We pass the original claim along with the unified retrieval string formed in the previous step as input, exactly as shown in Appendix \ref{ap:pt}; However, due to the limited context window of the LLM we are using, errors may arise. In such cases, we reduce the number of documents in the unified retrieval string and prompt the LLM again with a shorter input length. The reason we include only the retrievals in the unified retrieval string and omit the key points is that we want to avoid influencing the evidence construction process—specifically, the creation of question-answer pairs—in our strategy. We aim to keep this process dynamic, based on the available selected knowledge and the claim’s purpose.

Since the number of adequate question-answer pairs available as evidence for any claim may vary, we limit the LLM to providing at most 4 pairs to avoid penalties from additional, non-essential question-answer pairs in our prompt. The justification is needed to reason about the verdict based on the evidence and to directly write the predicted verdict afterward. Since the task requires the URL and scraped text for each item of evidence, we instruct the LLM to provide the citation ID when answering questions. This ensures that we can show the source for our verdict and each question-answer pair.

\section{Experiments and Results}

\subsection{Experimental Setup}
In this work, we utilized the \textit{Llama-3-70B model} for both steps described in Sections \ref{firsts} and \ref{thirds}, using the Groq API\footnote{https://groq.com/}. Additionally, we set the \textit{temperature} to 0 to ensure reproducibility and \textit{top}\_\textit{p} to 0.8. For key point construction, we set \textit{max}\_\textit{length} to 512, and for zero-shot prediction, we set it to 1024. In the retrieval step using BM25, we set \textit{top}\_\textit{k} to 70 for the original claim and to 12 for other queries, which include key points from both normal and combined forms. For zero-shot prediction, which is the third step of the strategy, if a rate limit occurs due to input length limitations, we retain only the first 55 documents for the original claim and 9 documents for key point retrievals.

\subsection{Evaluation Metrics}
The AVeriTeC scoring follows a similar approach to FEVER \cite{thorne2018fever} and considers the correctness of the verdict label conditioned on the correctness of the evidence retrieved. The label will only be considered correct if it mathches with the gold label and the Hungarian meteor score between the predicted evidence and the gold evidence is at least 0.25. However, Unlike in FEVER using a closed source of evidence such as Wikipedia, AVERITEC is intended for use with evidence retrieved from the open web. Since the same evidence may be found in different sources, we cannot rely on exact matching to score retrieved evidence. As such, the shared task evaluation strategy instead rely on approximate matching. Specifically, the Hungarian Algorithm \cite{kuhn1955hungarian} is used to find an optimal matching of provided evidence to annotated evidence.

\subsection{Main Results}
Despite our framework's straightforward procedure, which does not require any fine-tuning and only utilizes ZSL, as depicted in Table \ref{table:results}, it robustly improves the baseline in both retrieval scores—calculated for questions alone and for questions with answers—and the AVeriTeC score. This includes improvements of 0.14, 0.04, and 0.16 in retrieval scores for questions only, retrieval scores for questions with answers, and the AVeriTeC score, respectively. Based on these results, by using an open-source LLM, our framework has achieved a 10\textsuperscript{th} rank among all 23 system result submissions.

\section{Conclusion}
In this paper, we introduced ZSL-KeP, an effective yet straightforward framework for automated fact-checking. We utilized the ZSL capability of LLMs and constructed key points for extensive retrieval to generate evidence in a question-and-answer pairs format, along with a final verdict. By relying solely on the ICL capability of LLMs, our strategy operates without requiring any fine-tuning and is more straightforward compared to the baseline. Our framework sets a new benchmark, indicating promising avenues for future research in related topics.

\section{Limitations}
While our work shows strong performance, it has some limitations that suggest areas for future research. Our method improves diversity by using zero-shot key points for retrieval, but the limited input length of our LLM, constrained by time and budget limitations, prevented us from retrieving a larger document set. Additionally, a more powerful LLM could enhance accuracy in generating evidence and verdicts. Addressing these issues could significantly improve our framework's results.

\bibliography{custom}

\appendix

\section{Prompt Templates}
\label{ap:pt}
\begin{figure*}
	\centering
	\includegraphics[width=6in]{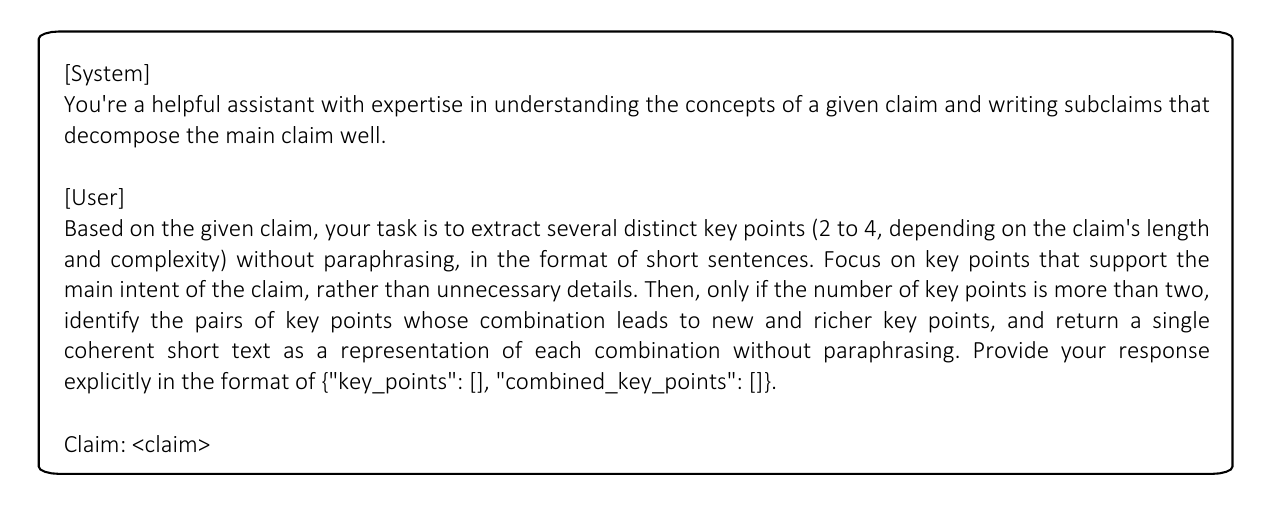}
	\caption{The Prompts for Zero-Shot Key Points Construction}
	\label{fig:fig2}
	
\end{figure*}

\begin{figure*}
	\centering
	\includegraphics[width=6in]{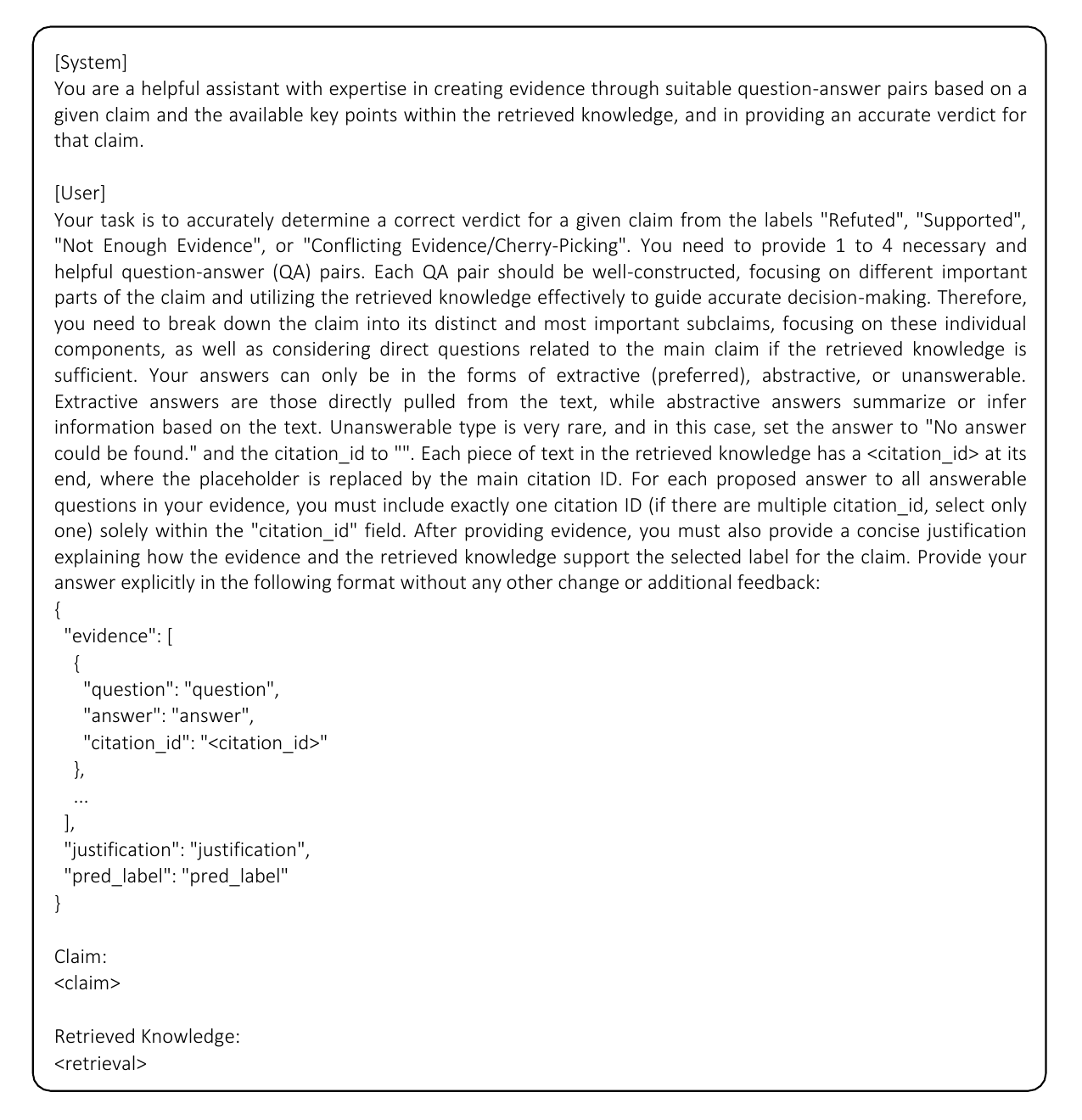}
	\caption{The Prompts for Zero-Shot Prediction}
	\label{fig:fig3}

\end{figure*}

This section provides all of the prompting templates used within the strategy. Figure \ref{fig:fig2} illustrates the full prompts for section \ref{firsts}, while the prompts for section \ref{thirds} are shown in Figure \ref{fig:fig3}. It is noteworthy that in the user messages, the tags “\textit{<claim>}” and “\textit{<retrieval>}” are replaced by the original claim and the unified retrieval string, respectively.

\end{document}